\documentclass[conference]{IEEEtran}
\IEEEoverridecommandlockouts
\usepackage{cite}
\usepackage{amsmath,amssymb,amsfonts}
\usepackage{bm}
\usepackage{tikz}
\usetikzlibrary{positioning, arrows}
\usepackage{algorithmic}
\usepackage{graphicx}
\usepackage{textcomp}
\usepackage{xcolor}
\usepackage{url}
\def\BibTeX{{\rm B\kern-.05em{\sc i\kern-.025em b}\kern-.08em
    T\kern-.1667em\lower.7ex\hbox{E}\kern-.125emX}}

\DeclareMathOperator*{\argmax}{arg\,max}
\begin{document}

\title{Latent Variable Double Gaussian Process Model for Decoding Complex Neural Data
\thanks{This project is sponsored by the Defense Advanced Research Projects Agency (DARPA) under cooperative agreement No. N660012324016. The content of the information does not necessarily reflect the position or the policy of the Government, and no official endorsement should be inferred.}
}

\author{\IEEEauthorblockN{Navid Ziaei}
\IEEEauthorblockA{
\textit{Department of Electrical} \\
\textit{\& Computer Eng.} \\
\textit{Isfahan University of Technology}\\
\textit{Department of Computer Science} \\
\textit{Worcester Polytechnique Institute}\\
Worcester, MA, USA \\
nziaei@wpi.edu}
\and
\IEEEauthorblockN{Joshua J. Stim}
\IEEEauthorblockA{
\textit{Department of Psychiatry} \\
\textit{\& Behavioral Sciences} \\
\textit{University of Minnesota}\\
Minneapolis, MN, USA \\
jstim1@jh.edu}
\and
\IEEEauthorblockN{Melanie D. Goodman-Keiser}
\IEEEauthorblockA{
\textit{Department of Psychiatry} \\
\textit{\& Behavioral Sciences} \\
\textit{University of Minnesota}\\
Minneapolis, MN, USA \\
mgoodman@umn.edu}
\and
\IEEEauthorblockN{Scott Sponheim}
\IEEEauthorblockA{
\textit{Department of Psychiatry} \\
\textit{\& Behavioral Sciences} \\
\textit{University of Minnesota}\\
\textit{Minneapolis Veterans Affairs }\\
\textit{Health Care System}\\
Minneapolis, MN, USA\\
sponh001@umn.edu}
\and
\IEEEauthorblockN{Alik S. Widge}
\IEEEauthorblockA{
\textit{Department of Psychiatry \& Behavioral Sciences} \\
\textit{University of Minnesota}\\
Minneapolis, MN, USA \\
awidge@umn.edu}
\and
\IEEEauthorblockN{Sasoun Krikorian}
\IEEEauthorblockA{\textit{Department of Computer Science} \\
\textit{Worcester Polytechnique Institute}\\
Worcester, MA, USA \\
smkrikorian@wpi.edu}
\and
\IEEEauthorblockN{Ali Yousefi}
\IEEEauthorblockA{\textit{Department of Computer Science} \\
\textit{Worcester Polytechnique Institute}\\
Worcester, MA, USA \\
ayousefi@wpi.edu}
}

\maketitle

\begin{abstract}
Non-parametric models, such as Gaussian
Processes (GP), show promising results in the analysis of complex
data. Their applications in neuroscience data have recently gained
traction. In this research, we introduce a novel neural decoder
model built upon GP models. The core idea is that two GPs
generate neural data and their associated labels using a set of low-
dimensional latent variables. Under this modeling assumption, the
latent variables represent the underlying manifold or essential
features present in the neural data. When GPs are trained, the
latent variable can be inferred from neural data to decode the
labels with a high accuracy. We demonstrate an application of this
decoder model in a verbal memory experiment dataset and show
that the decoder accuracy in predicting stimulus significantly
surpasses the state-of-the-art decoder models. The preceding
performance of this model highlights the importance of utilizing
non-parametric models in the analysis of neuroscience data.
\end{abstract}

\begin{IEEEkeywords}
Latent Variable Model, Gaussian Process Model, Dimensionality Reduction, Manifold Representation, Neural Decoder
\end{IEEEkeywords}

\section{Introduction}
With advancements in neural recording technology, we
observe consistent growth in the size and modality of neural
data being recorded. In parallel, machine learning (ML) tools
and techniques have shown unprecedented growth and
advancement in the last decade. This parallel growth holds the
promise for further discoveries of brain function. Specifically,
we anticipate the emergence of new neural decoder models that
achieve a higher level of prediction accuracy in inferring
underlying processes or factors influencing neural activity. For
example, we expect a more precise estimation of a rat's
movement trajectory by observing the spiking activity of an
ensemble of place cells, or decoding arm movement observing
motor cortex neural activity.

GPs are non-parametric models often used for both
regression and classification tasks \cite{Williams2006}. These models exhibit a
high expressive power; furthermore, they draw a more precise
picture of their estimations as they provide a measure of
uncertainty in their predictions. Extensions of GPs, such as the
latent variable GP, have been recently introduced for
dimensionality reduction, showing promising results in
inferring low-dimensional manifolds representing neural
activity \cite{Titsias2010}. While these methods open the door to better
characterizing neural data, they also have their limitations. For
instance, most of these approaches draw a point estimate in their
inference step, making them prone to overfitting. Additionally,
stimuli information is often discarded in the estimation of the
underlying manifold, limiting their decoding accuracy.

In this research, we develop a novel latent variable GP
model that characterizes both neural activity and associated
stimuli information \cite{Ziaei2024}. Two GPs, derived through a common
latent variable, characterize neural activity and associated
stimuli (or labels). For the model, we derive a Bayesian
inference solution allowing us to have a robust inference of the
underlying manifold. We show that the inferred manifold
captures essential features present in neural recordings. We
further show these features can be used in predicting the data
label or stimulus with high accuracy. The double GP and its inference solution resemble an auto-encoder framework, where
the input to the encoder is neural activity, and the decoder
generates both the label and neural data. However, the
fundamental difference lies in the framework's ability to run on
datasets with a limited sample size. With our model, we can find
the proper dimension of the manifold, whilst this is not the case
for the auto-encoder models. More importantly, our proposed
framework provides probabilistic inference and prediction,
leading to a more robust overall performance.

\section{Latent Varaible Double GP}
In this section, we first define the model. We then discuss
model training, inference, and finally the decoding processes.

\subsection{Model Definition}

Let’s assume we observe neural data as continuous
measurement defined by \(\mathbf{Y}^n \in \mathbb{R}^{N \times D}\), where \(N\) is the number of
data points and \(D\) represents the number of neural features. We
also have  \(\mathbf{Y}^s \in \mathbb{R}^{N \times K}\) , with \(K\) as the number of possible
categories or stimuli associated with each data point. We further
assume there is a latent variable \(\mathbf{X} \in \mathbb{R}^{N\times Q}\), where \(Q \ll D\), that captures underlying essential features deriving both \(\mathbf{Y}^n\) and \(\mathbf{Y}^s\).
Each row of \(\mathbf{X}\) is assumed to be generated from a multi-variate Gaussian distribution with mean \(\mathbf{0}\) and covariance \(\mathbf{I}_Q\) and these
latent variables are I.I.D; thus, their prior distribution is defined
by:
\begin{equation}
    p\left(\mathbf{X}\right) = \prod^N_{i=1} \mathcal{N}\left(\mathbf{x_i}; \mathbf{0}, \mathbf{I}_Q\right),
    \label{eq:first}
\end{equation}
where \(\mathbf{x}_i\) is the vector of the latent variable for the $i$th sample. We then assume that there are two GP prior on function values \(\mathbf{F}^n \in \mathbb{R}^{N\times D}\) and \(\mathbf{F}^s \in \mathbb{R}^{N \times K}\) defined by

\begin{align}
p\left(\mathbf{F}^n \mid \mathbf{X}\right) = \prod^D_{d=1} \mathcal{N}\left(f^n_{:, d}; \mathbf{0}, \mathbf{K}^n_{NN}\right)\\
p\left(\mathbf{F}^s \mid \mathbf{X}\right) = \prod^K_{k=1} \mathcal{N}\left(f^s_{:,k}; \mathbf{0}, \mathbf{K}^s_{NN}\right),
\end{align}
where \(\mathbf{K}^n_{NN}\) and \(\mathbf{K}^s_{NN}\) are the \(N \times N\) covariance matrices
defining functions behind the continuous and discrete
observations. \(\mathbf{f}^n_{:,d} \triangleq \left[f^n_d\left(\mathbf{x}_1\right),\ldots,f^n_d\left(\mathbf{x}_N\right)\right]\) and \(\mathbf{f}^s_{:,d} \triangleq \left[f^s_d\left(\mathbf{x}_1\right),\ldots,f^s_d\left(\mathbf{x}_N\right)\right]\) are function values representing the \(d\)th and \(k\)th columns of the functions. Elements of covariance matrices
are defined by a kernel function, with \(k^n_{i,j} = k_\phi \left(\mathbf{x}_i, \mathbf{x}_j\right)\) for
different pairs of \(i\) and \(j\) indices. Similarly, we have \(k^s_{i,j} = k_\theta \left(\mathbf{x}_i, \mathbf{x}_j\right)\) for \(\mathbf{K}^s_{NN}\) covariance. Here, \(\phi\) and \(\theta\) are the free parameters of the kernel function. We finally assume:
\begin{equation}
    y_{i,d} = f^n_d \left(\mathbf{x_i}\right) + \epsilon_d,
\end{equation}
where \(\epsilon_d \sim \mathcal{N}\left(0, \sigma^2_d\right)\) and \(\mathbf{F}^n = \left[\mathbf{f}^n_{:,1}, \ldots, \mathbf{f}^n_{:,D}\right]\) is the GP defined by \(\mathbf{K}^n_{NN}\). For the discrete process, we have 
\begin{equation}
    y_{i,k} = g\left(f^s_k\left(\mathbf{x}_i\right)\right),
    \label{eq:last}
\end{equation}
in which \(g\left(\cdot\right)\) is representing conditional probability of the \(y_{i,k}\)
defined by a squeezing function and \(\mathbf{F}^s = \left[\mathbf{f}^s_{:, 1}, \ldots, \mathbf{f}^s_{:, D}\right]\) is the GP defined by \(\mathbf{K}^s_{NN}\). 

\begin{figure}[t]
\centering
\begin{tikzpicture}[edge from parent/.style={draw, ->}]
    \node[draw, circle] at (0, 0) {\(\mathbf{X}\)}
    child {node[draw, circle] (Fn) {\(\mathbf{F}^n\)}
        child{node[draw, circle] (Yn) {\(\mathbf{Y}^n\)}}
    }
    child {node[draw, circle] (Fs) {\(\mathbf{F}^s\)}
        child{node[draw, circle] {\(\mathbf{Y}^s\)}}
    };
    \node[left=of Fn] (phi) {\(\phi\)};
    \draw[->] (phi) -- (Fn);
    
    \node[right=of Fs] (theta) {\(\theta\)};
    \draw[->] (theta) -- (Fs);

    \node[left=of Yn] (sigma) {\(\sigma\)};
    \draw[->] (sigma) -- (Yn);
\end{tikzpicture}
\caption{ Graphical Model Depicting Latent Variable Double GP: The latent variable $\mathbf{X}$ with a Gaussian prior derives both continuous and discrete covariance structures \(\mathbf{K}^n_{NN}\) and \(\mathbf{K}^s_{NN}\) for \(\mathbf{F}^n\) and \(\mathbf{F}^s\) value functions. These functions then define the generative model creating
the discrete and continuous measurements.
}
\label{fig:hierarchical-GP}
\end{figure}
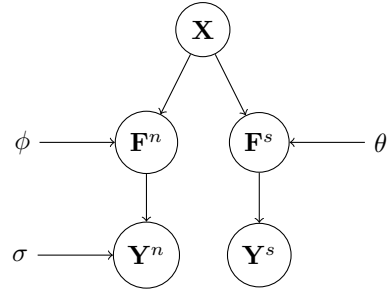

Eq.~\eqref{eq:first} to~\eqref{eq:last} define the generative model describing
how samples of continuous and discrete observations are created
as a function of the latent variables. Model parameters are tuned
to enhance the likelihood of observed data, which will be
elaborated in the following subsection on the training algorithm.

\subsection{Model Training}
For the training, we require to find the maximum
likelihood estimate of kernel parameters \(\phi\), \(\theta\), and measurement noise \(\sigma_d\). In our model, \(\mathbf{X}\) is unobserved; if \(\mathbf{X}\) were known, the conditional distribution of observation could be defined by:
\begin{align}
    \setlength{\jot}{-2pt}
    &\textstyle p\left(\mathbf{Y}^n,\mathbf{Y}^s \mid \mathbf{F}^n, \mathbf{F}^s, \mathbf{X}\right) = 
    p\left(\mathbf{Y}^n \mid \mathbf{F}^n, \mathbf{X}\right) p\left(\mathbf{Y}^s \mid \mathbf{F}^s, \mathbf{X}\right) \notag \\
    &p\left(\mathbf{Y}^n \mid \mathbf{F}^n, \mathbf{X}\right) = 
    \textstyle \prod_{i=1}^N \textstyle \prod_{d=1}^D 
        p\left(y_{i,d} \mid f^n_d\left(\mathbf{x}_i\right), \mathbf{X}\right)\\
    &p\left(\mathbf{Y}^s \mid \mathbf{F}^s, \mathbf{X}\right) = \textstyle \prod_{i=1}^N \textstyle \prod_{k=1}^K p\left(y_{i,k} \mid f^s_k \left(\mathbf{x}_i\right), \mathbf{X}\right)
\end{align}
The evidence for \(\mathbf{Y}^n\), \(\mathbf{Y}^s\) is defined by:
\begin{equation}
    p\left(\mathbf{Y}^n, \mathbf{Y}^s\right) = \int p\left(\mathbf{Y}^n, \mathbf{Y}^s \mid \mathbf{F}^n, \mathbf{F}^s, \mathbf{X}\right)p\left(\mathbf{X}\right) \, d\mathbf{F}^n \, d\mathbf{F}^s \, d\mathbf{X},
\end{equation}
which needs to be maximized by adjusting \(\phi\) and \(\theta\). Solving this
integral is an interactable optimization problem; instead, we use
doubly stochastic variational inference method introduced in \cite{Salimbeni2017} to estimate the parameters. In this approach, we introduce \(\mathbf{Z}^n\) and \(\mathbf{Z}^s\) as sets of inducing points along with corresponding inducing variables \(\mathbf{U}^n\) and \(\mathbf{U}^s\), each assumed to have a Gaussian Process (GP) prior. The goal is to draw an approximate posterior
distribution over latent features \(p\left(\mathbf{X} \mid \mathbf{Y}^n, \mathbf{Y}^s \right)\) using a variational distribution \(q_\psi \left(\mathbf{X}\right)\). To find this approximate posterior, we approximate \(p\left(\mathbf{F}^n, \mathbf{U}^n \mid \mathbf{Y}^n, \mathbf{X}\right)\) with \(q\left(\mathbf{F}^n, \mathbf{U}^n\right) = p\left(\mathbf{F}^n \mid \mathbf{U}^n, \mathbf{X}\right) q_\lambda \left(\mathbf{U}^n\right)\), and \(p\left(\mathbf{F}^s, \mathbf{U}^s \mid \mathbf{Y}^s, \mathbf{X}\right)\) with \(q\left(\mathbf{F}^s, \mathbf{U}^s \right) = p\left(\mathbf{F}^s \mid \mathbf{U}^s, \mathbf{X}\right) q_\gamma \left(\mathbf{U}^s\right)\). In this framework, \(q_\psi \left(\mathbf{X}\right)\), \(q_\lambda\left(\mathbf{U}^n\right)\), and \(q_\lambda \left(\mathbf{U}^s\right)\) represent the variational distribution, facilitating the decomposition of the model's complexity into manageable components. These distributions are parameterized as follows:

\begin{align}
    q_\psi\left(\mathbf{X}\right) &= \textstyle \prod^N_{i=1} \mathcal{N}\left(\mathbf{x}_i \mid \bm{\mu}_i, \mathbf{s}_i I_Q\right) \\
    q_\lambda\left(\mathbf{U}^s\right) &= \textstyle \prod^K_{k=1} \mathcal{N}\left(\mathbf{u}^s_k \mid \mathbf{m}^s_k, \mathbf{S}^s_k\right) \\
    q_\gamma\left(\mathbf{U}^n\right) &= \textstyle \prod^K_{k=1} \mathcal{N}\left(\mathbf{u}^n_k \mid \mathbf{m}^n_k, \mathbf{S}^n_k\right),
\end{align}
where $\psi \triangleq \left\{\bm{\mu}_i, \mathbf{s}_i\right\}^N_{i=1}$, $\lambda \triangleq \left\{\mathbf{m}^s_k, \mathbf{S}^s_k \right\}^K_{k=1}$, and $\gamma \triangleq \left\{\mathbf{m}^n_k, \mathbf{S}^n_k \right\}^K_{k=1}$ are variational parameters that need to be
optimized. Utilizing these variational distributions and
parameters, we can derive a lower bound for the likelihood
function, known as the Evidence Lower Bound (ELBO), defined
by:
\begin{align}
    \mathrm{ELBO} = &\mathbb{E}_{q_\psi\left(\mathbf{X}\right)}\left[\mathbb{E}_{p\left(\mathbf{F}^s \mid \mathbf{U}^s, \mathbf{X}\right)q_\lambda\left(\mathbf{U}^s\right)} \left[\log p\left(\mathbf{Y}^s \mid \mathbf{F}^s\right)\right]\right] \notag \\
    + &\mathbb{E}_{q_\psi\left(\mathbf{X}\right)}\left[\mathbb{E}_{p\left(\mathbf{F}^s \mid \mathbf{U}^s, \mathbf{X}\right)q_\lambda\left(\mathbf{U}^s\right)} \left[\log p\left(\mathbf{Y}^s \mid \mathbf{F}^s\right)\right]\right] \notag \\
    - &\mathrm{KL}\left(q_\lambda \left(\mathbf{U}^s\right) \| p\left(\mathbf{U}^s\right)\right) - \mathrm{KL}\left(q_\lambda\left(\mathbf{U}^n\right) \| p\left(\mathbf{U}^n\right)\right) \notag \\
    - & \mathrm{KL}\left(q_\psi\left(\mathbf{X}\right) \| p\left(\mathbf{X}\right)\right) \notag \\
    \triangleq & \mathrm{ELL}^s + \mathrm{ELL}^n - \mathrm{KL}_{u^s} - \mathrm{KL}_{u^n} - \mathrm{KL}_X.
    \label{eq:ELL}
\end{align}
The Expected Log Likelihood (ELL) terms in the lower bound,
defined in equation~\eqref{eq:ELL}, can be calculated using sampling
method. The details of these calculations are presented in \cite{Ziaei2024}.
Using a re-parameterization technique, we use gradient-based
approaches such as Adam to optimize the model’s free
parameters along with variational parameters. In summary, we
build a recursive solution to estimate variational parameters (\(\psi\), \(\lambda\), and \(\gamma\)) and other model’s free parameters.

\subsection{Inference Process}
In the inference step, our objective is to represent one or
several test points in the latent space. These test points may be
neural data \(Y^n_{*}\), stimulus labels \(Y^s_*\), or both. For new points, we define a latent representation \(X_*\) and initialize its variational parameters randomly. Given a new test point, the posterior distribution \(p\left(\mathbf{X}_* \mid \mathbf{Y}^n_*, \mathbf{Y}^s_*, \mathbf{Y}^n, \mathbf{Y}^s\right)\) is approximated by \(q\left(\mathbf{X}_*\right) = \mathcal{N}\left(\mathbf{X}_* \mid \bm{\mu}_*, \mathbf{s}_* \mathbf{I}_Q\right)\). The optimization process
minimizes the negative lower bound with respect to the
variational parameters \(\bm{\mu}_*\) and \(\mathbf{s}_*\), while the inducing points and
other model variables are kept constant. This optimization is
expressed by:
\begin{equation}
    \argmax_{\bm{\mu}_*, \mathbf{s}_*} \left\{\sum_d \mathrm{ELL}^n_{*, d} + \sum_k \mathrm{ELL}^s_{*, k} - \mathrm{KL}\left(q_\psi\left(\mathbf{X}_*\right) \| p\left(\mathbf{X}_*\right)\right) \right\}.
\end{equation}
By solving this optimization problem, we obtain the estimate of
the posterior over the test points’ latent variables, i.e. mean and
variance.

\subsection{Decoding Process}
In the decoding step, we possess partial information about
the test points, the neural data \(\mathbf{Y}^n_*\), and our goal is to determine the probability distribution over the stimulus \(\mathbf{Y}^s_*\). Similar to the inference step, we define a corresponding latent variable for the test point. However, this time we optimize the following expression:
\begin{equation}
    \argmax_{\bm{\mu}_*, \mathbf{s}_*} \left\{\sum_d \mathrm{ELL}^n_{*,d} - \mathrm{KL}\left(q_\psi \left(\mathbf{X}_*\right) \| p\left(\mathbf{X}_*\right)\right)\right\}.
\end{equation}
given we lack the label data. By optimizing the function, we
determine the optimal values for the variational parameters of \(X_*\). Through the generative path for the label, we can find \(p\left(\mathbf{Y}^c_* | \mathbf{X}_* \right)\) and effectively decode labels.

\section{Neural Dataset}
In this study, we examine the verbal memory (VerbMem)
dataset recorded by Scott Sponheim et al. \cite{marquardt2022inefficient}. The verbal
memory task involves three key stages, known as encoding,
recall, and lexical decision blocks. During encoding, participants
judge the size of objects shown as words. In the recall stage, they
try to remember and recount as many words as possible. Finally,
in the lexical decision part, they distinguish between previously
seen words and new words, while their EEG data is recorded to
analyze verbal memory and recognition processes. The EEG
data was collected using a 128-channel device, and participants'
responses included both correct and incorrect choices, along
with their response times.

Our focus here is on the lexical decision blocks within the
experiment. The block consists of both new and old (seen in the
encoding phase) words. Participants must distinguish between
these two categories. The goal is to determine whether a
participant is responding to new or old words based on neural
activity recordings. On average, participants' decisions are
incorrect \(17.6 \pm 9.6\%\) of the time. Incorrect responses are
excluded as we are interested in instances where participants are
confident in their choices. We construct a decoder model for
each individual patient, who may have a diagnosis of
schizophrenia (SZ) or are non-psychiatric control participants.
In our decoding, we remain agnostic to this information. The
neural features include time-domain features like the mean
amplitudes of the N200, P300, and post-P300 components,
coherence features such as total coherency across frequencies,
coherence vectors in different brain regions, and frequency-
domain features like the average power in delta, theta, alpha,
beta, and gamma bands, totaling 1705 features per trial \cite{Hu2019}. We
selected the 25 most correlated features based on the point-
biserial correlation coefficient and used them as input to our
decoder model. In the next section, we'll explain how our neural
decoder model works.

\section{Decoding Results in VerbMem Dataset}
\begin{figure}[t] 
    \centering
    \includegraphics[width=\columnwidth]{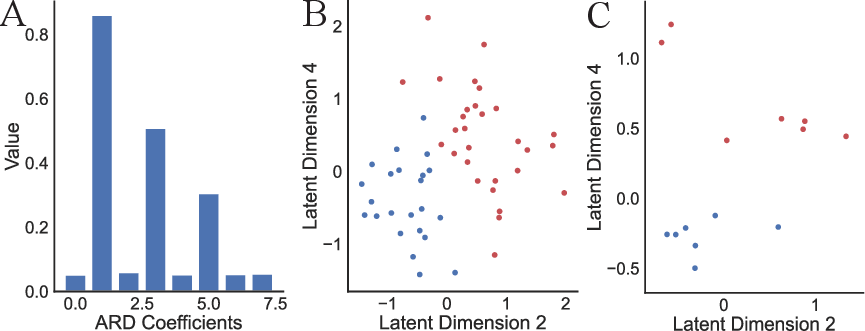} 
    \caption{Neural Features and Inferred $\mathbf{X}$: (A) ARD coefficients, indicating the relevance of each dimension in the input data to the model. (B-C) Scatter plots of latent variables for training (B) and testing (C) phases. Blue dots represent trials with new words, and red dots indicate trials with old words.}
    \label{fig2} 
\end{figure}

In our model---and for classification purposes---we assume the latent variable dimension is 8. For the kernels, we use Gaussian Automatic Relevance Determination (ARD) kernels. The $\phi$ and $\theta$ parameters include length scale and variance, which are estimated along with the approximate posterior distributions of $\mathbf{X}$. We aggregate three blocks of data, yielding approximately 120 data points per participant. Following the methodology outlined in Section III, trials with incorrect decisions are excluded, resulting in 90 to 105 trials per participant. We employ 5-fold cross-validation to evaluate decoding accuracy. After optimizing the model parameters on the train set, the ARD kernel parameter, as illustrated in Fig.~\ref{fig2}.A, reveals three dominant dimensions that significantly impact the model's predictions. This finding indicates that the model automatically identifies the necessary dimensions for the latent space. Fig.~\ref{fig2}.B illustrates the inferred $\mathbf{X}$ for training data, influenced by both neural features (continuous path) and labels (discrete path). The test set $\mathbf{X}_*$ is derived solely from neural features (see Fig.~\ref{fig2}.C). Then it goes through the discrete path of the generative model to determine the probability of trials being old or new. This discrimination yields a test set accuracy of 92\%.

\begin{table}[t]
    \centering
    \caption{Classification Accuracy in VERBMEM Dataset}
    \begin{tabular}{c cc cc }
    \hline
    & \multicolumn{2}{c}{XGBoost} & \multicolumn{2}{c}{Proposed model} \\
    \hline
    & Accuracy & F-measure & Accuracy & F-measure \\
    \hline
    1 & $0.71 \pm 0.11$ & $0.72 \pm 0.12$ & $\mathbf{0.85 \pm 0.05}$ & $\mathbf{0.86 \pm 0.06}$ \\
    2 & $0.77 \pm 0.15$ & $0.77 \pm 0.14$ & $\mathbf{0.82 \pm 0.05}$ & $\mathbf{0.82 \pm 0.06}$ \\
    3 & $\mathbf{0.74 \pm 0.14}$ & $\mathbf{0.74 \pm 0.15}$ & $0.72 \pm 0.11$ & $0.71 \pm 0.11$ \\
    4 & $0.73 \pm 0.08$ & $0.72 \pm 0.08$ & $\mathbf{0.75 \pm 0.07}$ & $\mathbf{0.75 \pm 0.07}$ \\
    5 & $0.67 \pm 0.04$ & $0.68 \pm 0.04$ & $\mathbf{0.68 \pm 0.06}$ & $\mathbf{0.69 \pm 0.06}$ \\
    \hline
    \end{tabular}
    \label{tab:verbmem-accuracy}
\end{table}

The training and decoding procedures were applied to data from five participants; the mean and variance of their classification accuracy from 5-fold cross-validation are detailed in Table~\ref{tab:verbmem-accuracy} The model shows consistent performance across participants, with an average accuracy of 76.4\%. To compare the decoder performance with state-of-the-art approaches, we examined the XGBoost classifier~\cite{chen2016xgboost}. As shown in Table~\ref{tab:verbmem-accuracy}, the model performance is significantly better than that of the XGBoost, and this holds for 4 out of 5 participants' data. The decoder's performance achieves an average F-score of 76.6\% across five participants, surpassing the XGBoost's F-score of 72.6\% by 4\%.

\section{Discussion}
In this research, we introduced a novel non-parametric neural decoder model. The model combines GP with latent variables, letting us simultaneously characterize neural activity and associated stimuli – or labels – as a function of the latent variables. The framework is well suited in analysis of datasets with limited sample sizes such as those datasets that are recorded in neuroscience experiments. We examined the model performance in decoding stimulus labels in the VerbMem dataset, reaching an average accuracy of 76.6\%. The performance is not the sole benefit of the model; with the model, we can also find the optimal dimension of latent variables. Furthermore, we can reduce the number of training samples to reduce the computational complexity of the decoding process without a loss of accuracy. We applied the XGBoost classifier to the same dataset; the XGBoost average prediction accuracy is 72.6\% which is about 4\% below our proposed approach. The preceding performance and other model attributes highlight its potential for high-dimensional neural data analysis. 
Whilst we listed the pros of the model, the model is not without its limitations. The model solution including inference, training, and decoding is not as straightforward as XGBoost and other ML techniques; this might hinder its wide usage in the neuroscience community. To address this, we have created an open-source package with synthetic and benchmark dataset examples, which we think will facilitate the usage of the framework in other datasets. The source code and additional resources are publicly available at our GitHub repository~\footnote{https://github.com/Navid-Ziaei/LDGD}. In this research, we mainly focused on decoding accuracy; we think it is crucial to study whether inferred latent variables have ethological interpretation and how they can be studied from a biomarker discovery point of view. A challenge with the GP model is its computational complexity when the size of the dataset grows; whilst this was not the case in the VerbMem dataset, it is important to examine how the model performance and computational cost scale with the sample size and dimension of the input. These aspects of the model present new avenues for future research, notably in exploring the model's effectiveness in clinical settings and its comparison with a broader range of ML techniques. Addressing these questions will be crucial in understanding the full potential and limitations of this innovative approach in the field of neuroscience.
The preceding performance of our latent variable Gaussian Process (GP) model in neural data analysis opens an intriguing pathway for future research in the realm of neural data analysis and neural decoder models. One of the crucial directions in our research is extending the model to handle time series data. This extension could significantly enhance the model's applicability, as we can detour feature extraction and pre-processing steps in the analysis of neural data and decoding step. In parallel, we are interested in addressing modeling challenges of latent GP models such as those that are listed in the discussion section.

\bibliographystyle{ieeetr}
\bibliography{main}

\end{document}